\PassOptionsToPackage{table}{xcolor}
\documentclass[11pt]{article}

\usepackage[preprint]{acl}

\usepackage{arydshln}
\usepackage{times}
\usepackage{amsmath}
\usepackage{amssymb}
\usepackage[table]{xcolor}
\usepackage{latexsym}
\usepackage{helvet}  
\usepackage{courier}  
\usepackage{graphicx} 
\usepackage{booktabs}
\usepackage{subcaption}
\usepackage{multirow}
\usepackage{natbib}  
\usepackage{caption} 

\usepackage{comment}

\usepackage[T1]{fontenc}

\usepackage[utf8]{inputenc}

\usepackage{microtype}

\usepackage{inconsolata}

\usepackage{graphicx}

\usepackage[most]{tcolorbox}
\tcbset{
  takeawaysstyle/.style={
    colback=orange!20,        
    colframe=orange!80!black, 
    fonttitle=\bfseries,      
    coltitle=white,           
    sharp corners,            
    boxrule=0.8pt,            
    top=3pt, bottom=3pt, left=6pt, right=6pt, 
    boxed title style={
      colback=orange!80!black, 
      sharp corners,
      boxrule=0pt,
      left=4pt, right=4pt, top=1pt, bottom=1pt,
    },
    before upper={\linespread{1}\selectfont}, 
  }
}

\title{The Evolution of Thought: Tracking LLM Overthinking \\via Reasoning Dynamics Analysis}

\author{
  Zihao Wei\textsuperscript{1,2},
  Liang Pang\textsuperscript{1,$\dagger$},
  Jiahao Liu\textsuperscript{},
  Wenjie Shi \textsuperscript{}, 
  Jingcheng Deng\textsuperscript{1,2},\\
  \textbf{Shicheng Xu}\textsuperscript{1,2}, 
  \textbf{Zenghao Duan}\textsuperscript{1,2},
  \textbf{Fei Sun}\textsuperscript{1},
  \textbf{Huawei Shen}\textsuperscript{1},
  \textbf{Xueqi Cheng}\textsuperscript{1} \\
  \textsuperscript{1}State Key Laboratory of AI Safety,Institute of Computing Technology,\\
Chinese Academy of Sciences, Beijing, China \\
  \textsuperscript{2}University of Chinese Academy of Sciences, Beijing, China \\
  \texttt{\{weizihao22z, pangliang\}@ict.ac.cn}
}

\begin{document}
\maketitle

\begin{abstract}
Test-time scaling via explicit reasoning trajectories significantly boosts large language model (LLM) performance but often triggers overthinking. To explore this, we analyze reasoning through two lenses: \textbf{Reasoning Length Dynamics}, which reveals a compensatory trade-off between thinking and answer content length that eventually leads to thinking redundancy, and \textbf{Reasoning Semantic Dynamics}, which identifies semantic convergence and repetitive oscillations. These dynamics uncover an instance-specific Reasoning Completion Point (RCP), beyond which computation continues without further performance gain. 
Since the RCP varies across instances, we propose a Reasoning Completion Point Detector (RCPD), an inference-time early-exit method that identifies the RCP by monitoring the rank dynamics of termination tokens (e.g., \texttt{</think>}).
Across AIME and GPQA benchmarks using Qwen3 and DeepSeek-R1, RCPD reduces token usage by up to 44\% while preserving accuracy, offering a principled approach to efficient test-time scaling.
\end{abstract}

\section{Introduction}
Test-time scaling via explicit reasoning trajectories has significantly enhanced the problem-solving capabilities of Large Language Models (LLMs) \cite{DBLP:journals/corr/abs-2501-12948, yang2025qwen3technicalreport,DBLP:journals/corr/abs-2509-18883}. 
While longer trajectories are generally associated with improved performance, recent studies reveal that this relationship is not strictly linear. 
Instead, excessive reasoning often leads to diminishing returns or even performance degradation, a phenomenon widely referred to as overthinking~\cite{ghosal2025doesthinkinghelpmirage,DBLP:journals/corr/abs-2509-23392}.

\begin{figure}[t]
    \centering
   \includegraphics[width=0.9\linewidth]{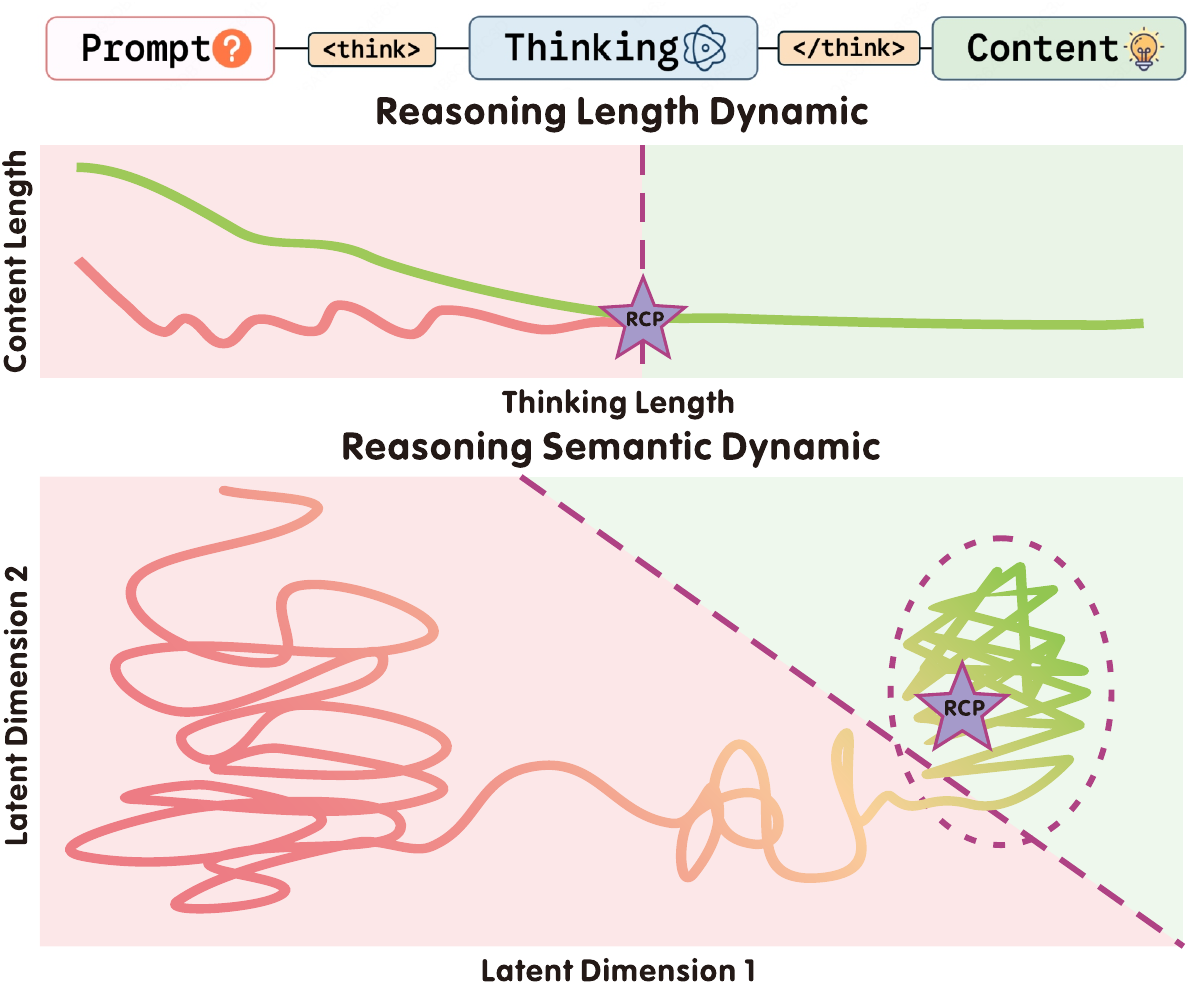}
    \caption{Overview of Reasoning Dynamics and RCP. The top panel summarizes Reasoning Length Dynamics, where content length decreases as thinking length grows under a thinking-content compensation regime until reaching the RCP. The bottom panel summarizes Reasoning Semantic Dynamics, where the latent semantic trajectory transitions from broad exploration to a stable neighborhood with repetitive oscillations, with the onset of convergence aligning with the RCP. The top and bottom panels are defined in \S\ref{sec:length_dynamics} and \S\ref{sec:semantic_dynamics}, respectively.}
    \label{fig:intro}
\end{figure}

This non-monotonic scaling behavior~\cite{wu2025lessunderstandingchainofthoughtlength} motivates a fundamental question: how does each step of the reasoning process incrementally influence the final outcome? Despite its importance, such a fine-grained analysis of the intermediate reasoning steps remains insufficiently explored. By quantifying the functional impact of each step on the final answer, we implement a study of reasoning dynamics to uncover the internal mechanisms of overthinking and identify the regime where further reasoning tokens become redundant. To explore this, we analyze the reasoning dynamic of LLM from two complementary perspectives. 

First, we explicitly characterize the inference as the interaction between the thinking process and the content process. Motivated by recent findings on the non-linear relationship between reasoning length and accuracy \cite{DBLP:journals/corr/abs-2505-00127,DBLP:journals/corr/abs-2502-07266}, we investigate their \textbf{Reasoning Length Dynamics} and uncover a phenomenon we term \textit{thinking-content compensation}, as shown in Figure~\ref{fig:intro}. Initially, the thinking process and content generation exhibit a compensatory relationship: as thinking trajectories expand, the subsequent content becomes increasingly concise. However, this dynamic eventually transitions into a saturation phase. In this regime, although the model has acquired sufficient information to formulate a correct answer, the thinking process continues to lengthen while the content remains unchanged, leading to redundancy.

Second, to uncover the underlying mechanism driving these surface-level length patterns, we further analyze the trajectory of high-dimensional representations in the semantic space~\cite{DBLP:journals/corr/abs-2509-23024}. In terms of \textbf{Reasoning Semantic Dynamics}, we identify \textit{semantic path convergence}, as shown in Figure~\ref{fig:intro}. We observe that these representations initially exhibit significant fluctuations, reflecting active exploration of the solution space. Subsequently, the inference path enters a convergence regime where the semantic representations converge toward a stable neighborhood, exhibiting repetitive oscillations within that localized space. This suggests that the solution hypothesis has stabilized.

Bridging these two dynamics suggests a structural explanation for overthinking: it arises when the model continues to generate reasoning tokens after the process has converged. This motivates a principled boundary between necessary reasoning and redundant computation after convergence. We call this boundary the Reasoning Completion Point (RCP). At the RCP, thinking-content compensation ends and the semantic trajectory begins to converge. This boundary is latent and instance-dependent, so mitigating overthinking requires detecting the RCP online during decoding rather than relying on a fixed reasoning length budget. Based on this formalization, we propose a Reasoning Completion Point Detector (RCPD), which monitors the rank of the special token designed to terminate the thinking process (e.g., \texttt{</think>}) to detect the RCP during decoding without additional computational overhead. Detecting the RCP enables early truncation of redundant reasoning, reducing token consumption while maintaining accuracy.

\section{Reasoning Length Dynamics}
\label{sec:length_dynamics}
Many reasoning-specialized LLMs expose an intermediate thinking process delimited by special tokens (e.g., \texttt{<think>} and \texttt{</think>}) and then transition to a content process that produces the final answer~\cite{DBLP:journals/corr/abs-2501-12948,yang2025qwen3technicalreport}. This section asks a simple but diagnostic question: as we allocate more steps to the thinking process, how do the resulting content length and answer correctness change? We study this via an intervention that forces the model to stop thinking at controlled steps and immediately begin generating content.

\subsection{Dynamics Setup: Stepwise Truncation}
\label{subsec:length_setup}

We formalize LLM generation as a sequential coupling of two distinct phases: a \textit{thinking process} $t$ and a \textit{content process} $y$. Even when the raw trajectory $t$ is withheld in proprietary models due to safety or competitive considerations~\cite{DBLP:journals/corr/abs-2412-16720,gemini3pro2025}, it remains the indispensable structural precursor that conditions the final output. Formally, the generation follows the joint distribution $P(y, t|x) = P(t|x)P(y|x, t)$, where $x$ denotes the input prompt.

To analyze the reasoning process at a granular level, we decompose both the thinking trajectory $t$ and the content trajectory $y$ into discrete \textit{reasoning steps} $s$. For a trajectory consisting of raw tokens, we segment it into steps, $t = (s_1, \dots, s_{K})$, where each step $s_k$ represents a complete sentence segmented via \texttt{NLTK}~\citep{DBLP:conf/acl/BirdL04}. This mapping allows us to intervene on the thinking budget by defining the thinking prefix up to step $k$ as $t_{1:k} = (s_1, \dots, s_k)$.

To explore how each step incrementally influences the outcome, we implement a \textbf{stepwise truncation protocol}. For each $k \in \{1, \dots, K\}$, we truncate the thinking trajectory immediately after $s_k$ and force-inject the end-of-thinking delimiter (e.g., \texttt{</think>}). This injected delimiter serves as an explicit control signal that terminates the internal thinking process and triggers the transition to content generation:
\begin{equation} \label{eq:stepwise_trunction}
    y_{k} \sim P(y \mid x, t_{1:k}).
\end{equation}

Under this protocol, we monitor two core length measures quantified by reasoning steps:
(1)~\textbf{Thinking Length ($L_t$):} The cumulative number of allowed reasoning steps in the thinking phase, $L_t=k$; (2)~\textbf{Content Length ($L_c$):} The number of steps in the content process $y^{(k)} = (s^y_1, \dots, s^y_{L_c})$, where $L_c$ is the count of steps generated in response to the truncated thinking prefix $t_{1:k}$.

By monitoring the evolution of $L_t$, $L_c$, and the corresponding answer correctness across $k$, we implement a dynamical analysis of the reasoning process and its transition toward redundancy.

\begin{figure*}[t]
  \centering
  \begin{subfigure}[t]{0.48\textwidth}
    \centering
    \includegraphics[width=1.0\linewidth]{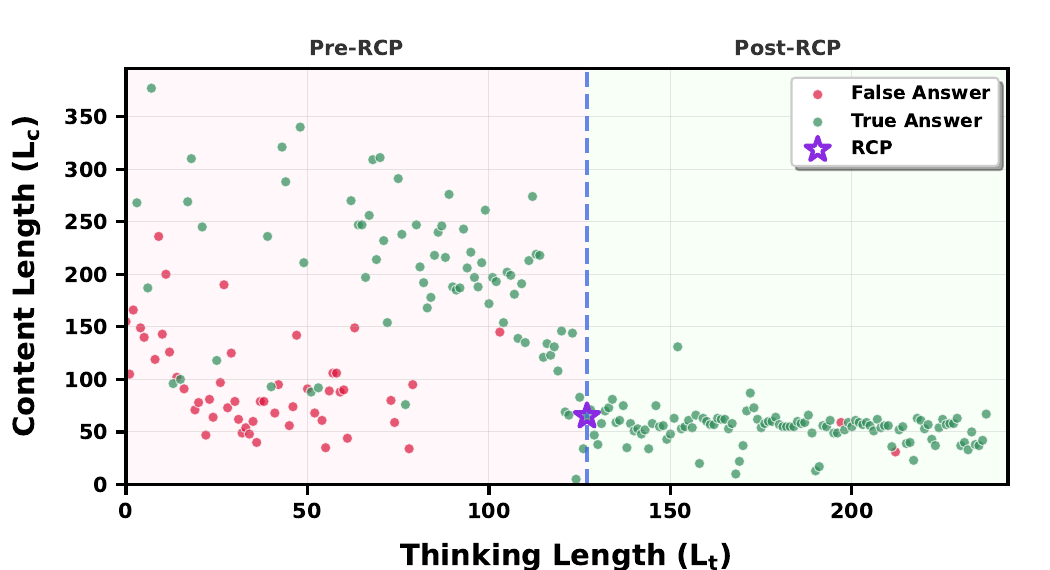}
    \caption{AIME24 Question 19 Length Dynamics}
    \label{fig:subA}
  \end{subfigure}
  \hfill
  \begin{subfigure}[t]{0.48\textwidth}
    \centering
    \includegraphics[width=1.0\linewidth]{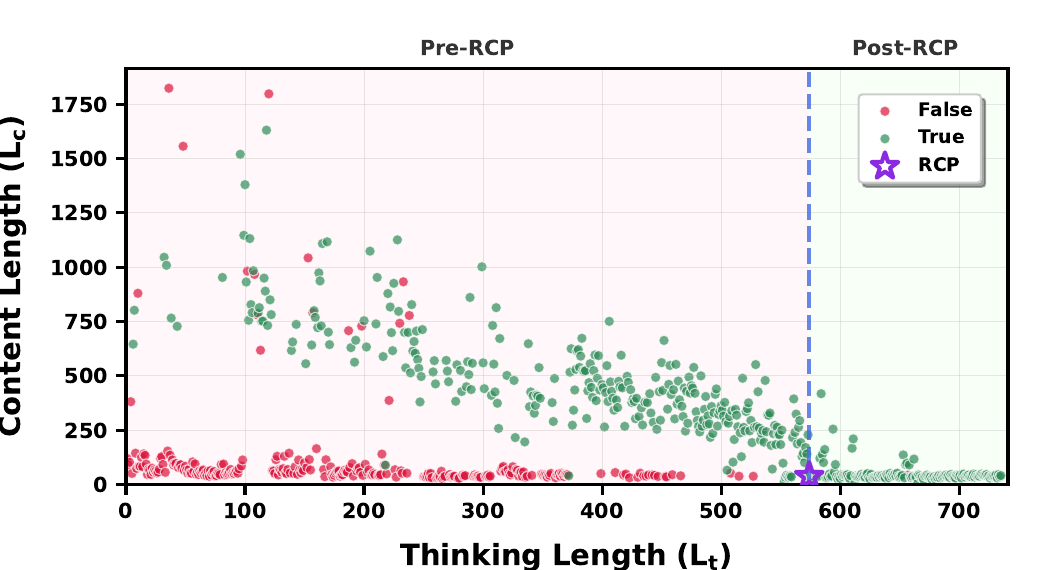}
    \caption{AIME25 Question 17 Length Dynamics}
    \label{fig:subB}
  \end{subfigure}
  \caption{Two-stage reasoning dynamics separated by RCP: the Pre-RCP Active Reasoning Stage and the Post-RCP Converged Reasoning Stage. The vertical dashed line indicates the RCP boundary. Additional examples are provided in Appendix Figure~\ref{fig:appendix_three_stage_additional}.}
  \label{fig:two_three_stage}
\end{figure*}

\subsection{Observations}
\label{subsec:length_observations}

Figure~\ref{fig:two_three_stage} provides a representative visualization of thinking-content compensation and the eventual stabilization pattern.

\begin{tcolorbox}[takeawaysstyle, title=Observation 1: Reasoning Compensation]
\textit{In the early stage, shorter thinking force the model to offload more reasoning into the content to maintain performance, resulting in an increased content length.}
\end{tcolorbox}

In the early region of Figure~\ref{fig:two_three_stage}, this effect appears directly as a separation between correct and incorrect samples. In Figure~\ref{fig:subA} around $k\!\approx\!50$, many incorrect answers cluster at very short content, whereas correct answers at the same $k$ often require substantially longer content. Figure~\ref{fig:subB} exhibits the same pattern: in the early region, incorrect answers concentrate near the bottom band, while correct answers extend to much larger content steps. These examples show that when the thinking budget is insufficient, maintaining correctness frequently requires expanding the content, and failures often manifest as terse outputs.

\begin{tcolorbox}[takeawaysstyle, title=Observation 2: Thinking Redundancy]
\textit{With sufficient thinking, content retracts to its conciseness and accuracy stabilizes; any further ineffective reasoning is manifested solely as an expansion of thinking.}
\end{tcolorbox}

In the late region of Figure~\ref{fig:two_three_stage}, content length collapses into a narrow band and stabilizes after the dashed line, remaining constant even as $k$ increases by over 100 steps (Figures~\ref{fig:subA}, \ref{fig:subB}). This confirms that additional thinking budget in this stage no longer alters content volume but primarily accumulates as redundancy.

\section{Reasoning Semantic Dynamics}
\label{sec:semantic_dynamics}

While Section~\ref{sec:length_dynamics} establishes a macroscopic trade-off between thinking and content length, these metrics are mere proxies that do not distinguish between active hypothesis refinement and futile repetition. To uncover the mechanistic trigger of overthinking, we shift from surface-level length statistics to fine-grained semantic-level evolution, namely Reasoning Semantic Dynamics.

Following the conceptualization of reasoning as a trajectory in latent semantic space~\cite{DBLP:journals/corr/abs-2509-21549}, we probe the model's internal evolution through its intermediate states. Since thinking tokens serve as intermediate computational steps, the most direct manifestation of the model's current state is the answer it produces when forced to terminate. By tracking the semantic trajectory across truncation steps, we can trace the stabilization of the model's emerging conclusion and map its search path.

Consequently, this section investigates a pivotal question: as the thinking process extends, how does the induced content evolve semantically, and when does it reach a state of terminal convergence?

\subsection{Dynamics Setup: Semantic Lens}
\label{subsec:sem_setup}

To explore the internal evolution of thoughts, we apply a semantic lens to the induced content distribution generated via the stepwise truncation protocol. For a fixed thinking prefix $t_{1:k}$, the semantics of a single content can be highly sensitive to decoding stochasticity. To mitigate this instability and obtain a robust representation of the model's hypothesis at each step $k$, we independently sample $M$ content, denoted as $\{y_k^{(i)}\}_{i=1}^M$ following the Eq.~\ref{eq:stepwise_trunction}.

Each continuation is mapped to a latent space using the Qwen3 Embedding model~\cite{DBLP:journals/corr/abs-2506-05176}, yielding vectors $v_k^{(i)}=\mathcal{E}(y_k^{(i)})\in\mathbb{R}^d$. To visualize these dynamics in a consistent coordinate system, we perform Principal Component Analysis (PCA) on the pooled set of embeddings over all truncation steps $k$ and samples $i$. We project each embedding to a 2D point $\tilde{v}_k^{(i)}\in\mathbb{R}^2$, and the resulting semantic trajectory is traced by the per-step centroid:
\begin{equation}
    \tilde{\mu}_k = \frac{1}{M} \sum_{i=1}^M \tilde{v}_k^{(i)}.
\end{equation}

The evolution of $\tilde{\mu}_k$ captures how the \emph{content semantics} mature as $L_t$ increases. We use the 2D view for interpretability.

\begin{figure}[t]
    \centering
    \includegraphics[width=\linewidth]{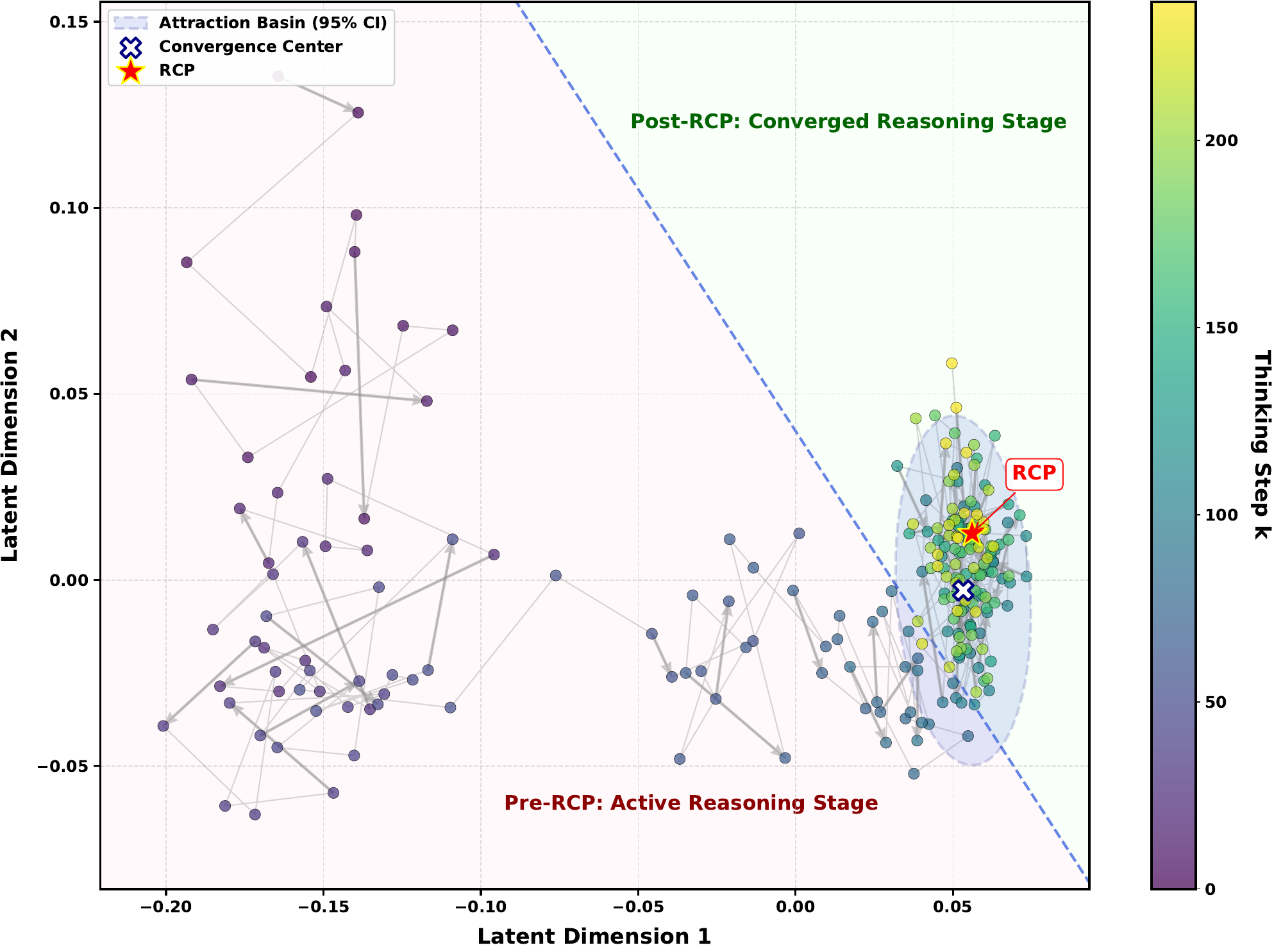}
    \caption{Semantic trajectory showing the transition from Pre-RCP Active Exploration to Post-RCP Reasoning Convergence. The dashed line indicates the RCP boundary. Additional examples are provided in Appendix Figure~\ref{fig:appendix_semantic_trajectory_additional}.}
    \label{fig:semantic_trajectory}
\end{figure}

\begin{figure}[t] 
    \centering
    \includegraphics[width=\linewidth]{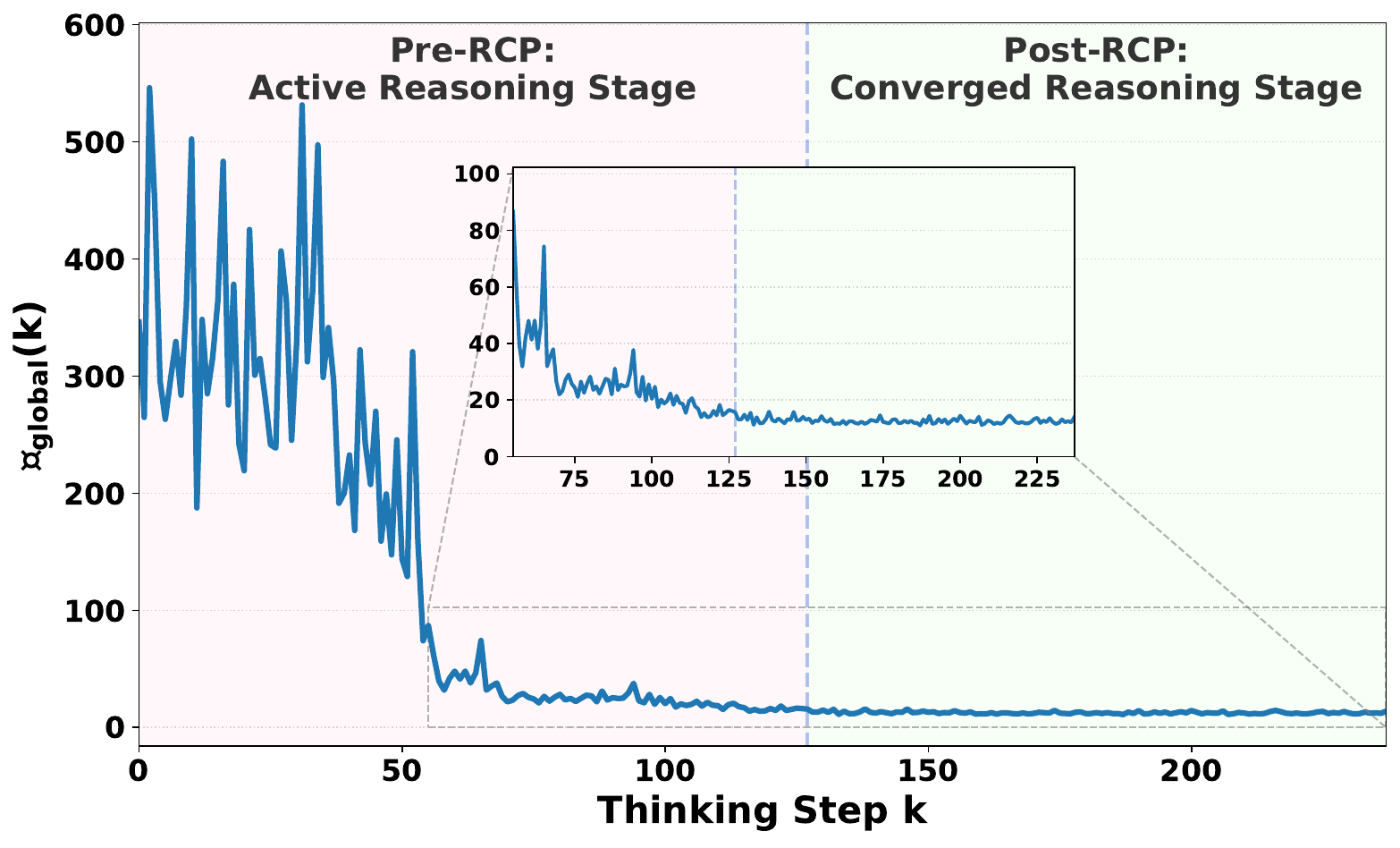}
    \caption{Semantic convergence residual over thinking steps. $\mathcal{D}_{\text{global}}(k)$ declines and then approaches a low plateau. The vertical dashed line indicates the RCP boundary; the inset zooms into the late-step region for readability.}
    \label{fig:reasoning_dynamics}
\end{figure}

\paragraph{High-Dimensional Convergence Metric.}
The 2D PCA projection is only for visualization. For quantitative analysis, we operationalize convergence via a semantic proxy distribution in a higher-dimensional PCA space that retains substantially more information than the 2D view.
We project embeddings to an $r$-dimensional PCA space ($r\gg2$), obtain $z_k^{(i)}\in\mathbb{R}^r$, and approximate the induced distribution by a Gaussian $Q_k=\mathcal{N}(\boldsymbol{\mu}_k,\Sigma_k)$:
\begin{equation}
\begin{aligned}
\boldsymbol{\mu}_k &= \frac{1}{N}\sum_{i=1}^N z_k^{(i)}, \\
\Sigma_k &= \mathrm{Cov}\!\left(\{z_k^{(i)}\}_{i=1}^N\right) + \lambda I,
\end{aligned}
\end{equation}
where $\lambda I$ is a small ridge term for numerical stability.
We estimate a terminal reference $Q_\infty$ by pooling samples from a short tail window of the final truncation steps and computing its mean and covariance.
We then measure the distance to the converged semantic regime by the global convergence residual
\begin{equation}
    \mathcal{D}_{\text{global}}(k)\triangleq D_{\text{KL}}(Q_k\parallel Q_\infty),
\end{equation}
which will later serve as a key ingredient for formalizing convergence.

\subsection{Observations}
\label{subsec:drift_converge}
Figure~\ref{fig:semantic_trajectory} provides a representative semantic view of the length-level dynamics in \S\ref{sec:length_dynamics}; we refer to the resulting transition as \textit{semantic path convergence}.

\begin{tcolorbox}[takeawaysstyle, title=Observation 3: Semantic Exploration]
\textit{During the initial phase, the semantic within the reasoning process undergo rapid shifts, indicating that the model is actively exploring feasible solution paths.}
\end{tcolorbox}

In the left region of Figure~\ref{fig:semantic_trajectory}, the projected induced samples are sparsely scattered, with large gaps between neighboring points, and the per-step mean $\tilde{\mu}_k$ makes large jumps across truncation steps, suggesting that the model is still actively revising its semantic hypothesis. 
Figure~\ref{fig:reasoning_dynamics} then quantifies this instability: $\mathcal{D}_{\text{global}}(k)$ varies widely in the early stage, indicating that the induced content distribution can shift markedly between adjacent truncation steps. This semantic instability also explains the length-level outcome bifurcation in Figure~\ref{fig:two_three_stage}: at the same moderate $k$, some runs compensate by generating longer content and reach the correct hypothesis, while others collapse to short, incorrect outputs.

\begin{tcolorbox}[takeawaysstyle, title=Observation 4: Reasoning Convergence]
\textit{As reasoning progresses, the semantics gradually converge toward a specific neighborhood, exhibiting repetitive oscillations within that localized space.}
\end{tcolorbox}

The right region in Figure~\ref{fig:semantic_trajectory} makes the convergence picture explicit: once the trajectory crosses the dashed line, the induced samples concentrate inside the 95\% attraction basin around the convergence center, and $\tilde{\mu}_k$ largely stays within that localized area. To make this stable-neighborhood claim testable, Appendix~\ref{sec:appendix_semantic} fits a 95\% confidence ellipse in the 2D PCA plane and finds that 97.3\% of post-transition points remain inside the basin.

Figure~\ref{fig:reasoning_dynamics} provides the corresponding high-dimensional signal: after the turning point marked by the dashed line, $\mathcal{D}_{\text{global}}(k)$ approaches a low plateau, indicating that the induced content distribution is already close to its terminal state. Notably, this non-zero floor reflects an intrinsic entropy gap between the instantaneous state $Q_k$ and the aggregate convergence basin $Q_\infty$, suggesting a dynamic equilibrium within the basin rather than collapse to a single singularity. This also aligns with the right region in Figure~\ref{fig:two_three_stage}, where thinking continues to grow while the content length stays almost unchanged. In this regime, extra steps mainly accumulate as redundant thinking and rarely alter the final answer.

These two observations together explain why more thinking can stop helping: once the content semantics have converged, additional thinking steps occur after the induced content distribution has effectively stabilized, making overthinking largely redundant.

Together with the length lens in \S\ref{sec:length_dynamics}, the semantic trajectory suggests a shared transition from an early stage of active semantic change to a late regime of semantic stationarity. 

\section{Reasoning Completion Point}
\label{sec:formalizing_dynamics}

The empirical observations in previous sections suggest that LLM reasoning is not a monolithic process but a bipartite evolution. By synthesizing thinking-content compensation and semantic path convergence, we identify a fundamental two-stage transition in the model's reasoning: \textbf{Pre-RCP: Active Reasoning} and \textbf{Post-RCP: Converged Reasoning}.

To formalize these stages, we first establish two quantitative metrics to capture the stability of the reasoning output. Consistent with the observations in \S~\ref{sec:length_dynamics}, as thinking length increases, the compensation effect eventually saturates. Content length $L_c$ ceases to change materially, and the induced content stabilizes. We formalize this step-to-step content change as:
\begin{equation}
\Delta_{\text{content}}(k)\triangleq \left|L_c(k)-L_c(k-1)\right|,
\end{equation}
where $L_c(k)$ is measured at truncation step $k$ following \S~\ref{subsec:length_setup}.
Content stabilization occurs when $\Delta_{\text{content}}(k)$ drops below a small threshold $\epsilon_c$ and remains near zero thereafter.

Simultaneously, building upon the semantic path analysis in \S~\ref{sec:semantic_dynamics}, the induced semantics transition from high-drift exploration to a stable neighborhood. We quantify this through the global convergence residual $\mathcal{D}_{\text{global}}(k)$, which stabilizes at a low plateau as thinking progresses. We operationalize the entry into this convergence region when $\mathcal{D}_{\text{global}}(k) \le \epsilon_D$, where $\epsilon_D$ is set by the empirical tail fluctuations used to estimate $Q_\infty$.

Based on these metrics, we categorize the reasoning process into two distinct stages:

\paragraph{Pre-RCP: Active Reasoning Stage.}
\begin{equation}
\label{eq:pre_rcp}
\mathcal{D}_{\text{global}}(k) > \epsilon_D \quad \lor \quad \Delta_{\text{content}}(k) > \epsilon_c .
\end{equation}
The model is in a state of high-drift exploration; the semantic path continues to evolve, and the induced content remains volatile, frequently expanding to compensate for insufficient thinking. Computation in this stage is typically essential for answer maturation.

\paragraph{Post-RCP: Converged Reasoning Stage.}
\begin{equation}
\label{eq:post_rcp}
\mathcal{D}_{\text{global}}(k) \le \epsilon_D \quad \land \quad \Delta_{\text{content}}(k) \le \epsilon_c .
\end{equation}
The semantic state stabilizes within a converged neighborhood, and the content length reaches a plateau. Additional compute beyond this point yields diminishing returns, often manifesting as redundant rationalization or "overthinking."

The critical boundary separating these two regimes is the \textbf{Reasoning Completion Point (RCP)}. Formally, we define the RCP as the earliest truncation step $k_{\text{RCP}}$ where both length stabilization and semantic convergence are simultaneously achieved:
\begin{equation}
    k_{\text{RCP}}\triangleq \min\left\{k: 
    \begin{aligned}
    &\Delta_{\text{content}}(k)\le \epsilon_c \\
    \land\  &\mathcal{D}_{\text{global}}(k)\le \epsilon_D
    \end{aligned}
    \right\},
    \label{eq:rcp_formal}
\end{equation}

The thresholds $\epsilon_c$ and $\epsilon_D$ are empirically determined by the tail fluctuations in the late-step window. Under this unified formalization, the RCP serves as an instance-specific "stopping criterion" that respects the intrinsic complexity of the prompt. Beyond $k_{\text{RCP}}$, additional thinking is largely redundant as it no longer materially alters the induced content's length or conceptual essence. Consequently, mitigating overthinking reduces to an online detection problem: identifying the transition $k_{\text{RCP}}$ during decoding to terminate the thinking phase at the precise moment of reasoning maturity.
\begin{figure*}[t]
    \centering
    \includegraphics[width=0.9\textwidth]{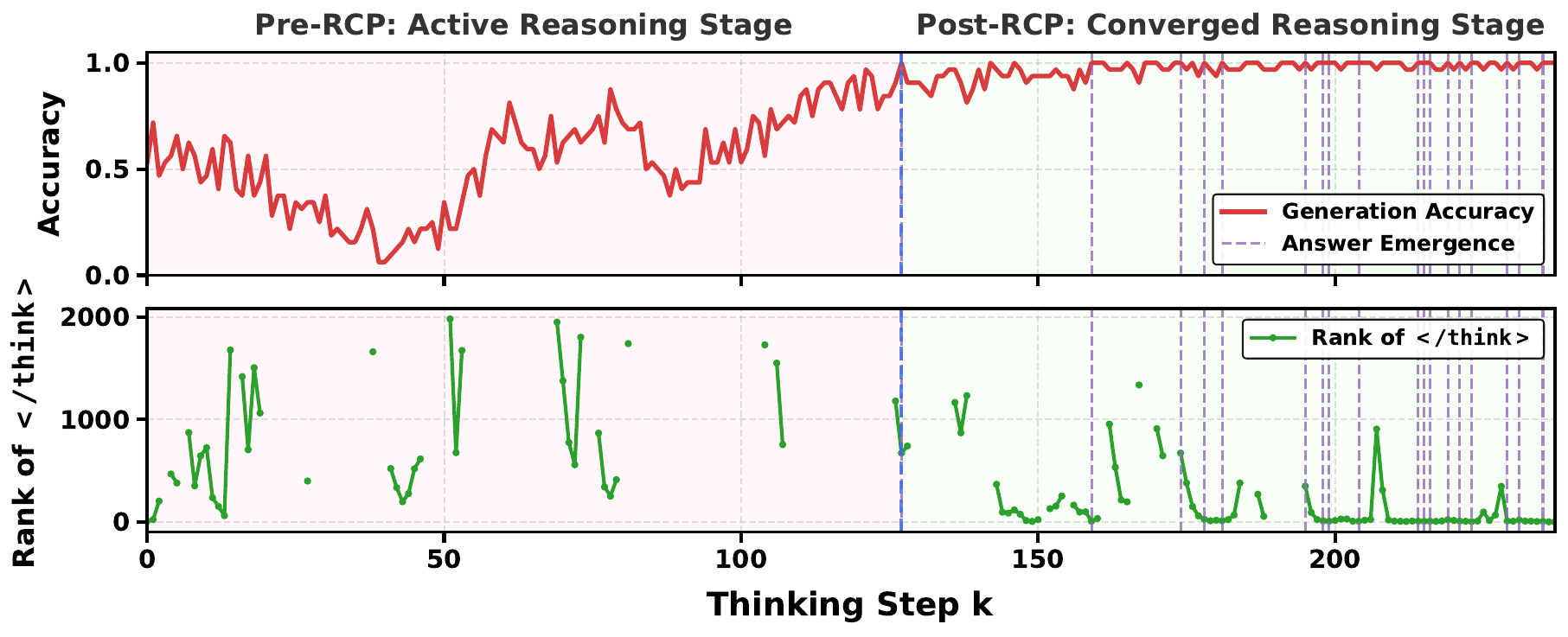}
    \caption{Top panel: Accuracy stabilizes around answer emergence. 
Bottom panel: The rank of \texttt{</think>} ($R_k$) drops precipitously at answer emergence; this drop serves as a signature of convergence.}
    \label{fig:reasoning_acc_rank}
\end{figure*}

\section{Online Overthinking Elimination}
The pivot to eliminating overthinking lies in the precise detection of the RCP. The logical progression of this section follows a "distill-to-detect" pipeline: we first leverage our reasoning dynamic findings (\S~\ref{sec:length_dynamics} and \S~\ref{sec:semantic_dynamics}) to extract gold RCP labels from full reasoning trajectories; however, to eliminate overthinking in practice, these insights must be translated into an online detector capable of real-time monitoring during inference. Specifically, we use the extracted offline data to train a tree-based model that identifies the latent transition into convergence. The resulting rules enable rapid RCP detection during standard decoding, effectively truncating redundant computation without sacrificing accuracy.

\subsection{Offline RCP Extraction}
In the ``distill'' stage of our pipeline, we start from the formal RCP criterion in Eq.~\ref{eq:rcp_formal} and extract instance-level annotations from full thinking trajectories. For each instance, we generate a complete reasoning trajectory, perform controlled truncation over candidate sentence boundaries to identify the earliest boundary at which reasoning has effectively converged, denoted $k_{\text{RCP}}$. This procedure is inherently offline, as it depends on access to complete reasoning trajectories and auxiliary semantic probes that are unavailable during standard decoding. Accordingly, we treat the resulting $k_{\text{RCP}}$ annotations as gold supervision.

A key empirical observation from these labeled traces is that $k_{\text{RCP}}$ typically coincides with the \textit{first emergence} of the final answer in the reasoning trajectories. 
While such answer emergence marks the onset of the Post-RCP convergence regime, it is not a reliable target for online detection. Answer surface forms vary widely across instances, including numbers, units, natural language, and multiple-choice letters, and may be rephrased, rendering string-based criteria brittle.
To replace this unobservable event with an intrinsic proxy, we examine the model's next-token distributions around $k_{\text{RCP}}$ and find a consistent transition in the rank $R_k$ of the thinking-termination delimiter \texttt{</think>}. Specifically, $R_k$ drops sharply as the answer first emerges, reflecting an increasing preference to terminate thinking. This delimiter-rank dynamics is internal to the model and largely invariant to answer format, making it the key distilled feature used for online RCP detection.

\subsection{Online RCP Detection}
Offline extraction reveals an online-accessible signature of convergence. When the final answer first emerges, the rank $R_k$ of the thinking-termination token \texttt{</think>} drops sharply (Figure~\ref{fig:reasoning_acc_rank}). This motivates a practical reformulation of online RCP detection: given the rank observations available up to the current sentence boundary, decide whether decoding has reached the Reasoning Completion Point and can stop.

We instantiate this idea as a lightweight Reasoning Completion Point Detector (RCPD). At each sentence boundary, RCPD queries the next-token distribution and extracts the current $R_k$ together with a short history of preceding rank values. Using offline-extracted $k_{\text{RCP}}$ as supervision on Math-500~\cite{lightman2023lets}, we label boundaries and fit an interpretable decision tree whose splits capture the implicit stopping criterion in terms of rank dynamics. We then distill the tree into deterministic stopping rules (exact thresholds and window sizes in Appendix~\ref{sec:appendix_stopping}).

During inference, we apply the same rules online; once triggered, we force \texttt{</think>} to terminate the thinking phase and immediately transition to answer generation. This completes the distill-to-detect pipeline and yields a practical stopping strategy for eliminating Post-RCP overthinking.

\subsection{Experimental Evaluation}
We evaluate on AIME24, AIME25~\cite{maa_aime}, and GPQA-D~\cite{rein2023gpqagraduatelevelgoogleproofqa}, using Qwen3 (8B/14B/30B-A3B) and DeepSeek-R1-Distill-8B. We report efficiency measured by the average number of generated tokens (Tok), Accuracy (Acc), and the relative performance-cost ratio (RPCR), which measures the accuracy gain over the No-Think baseline per generated token. For readability, we report RPCR with two decimal places. For ease of comparison, we show the Compression Rate (CR; relative to full decoding) as an integer percentage in parentheses next to Tok. Metric definitions and token accounting details are provided in Appendix~\ref{sec:appendix_eval_details}. For autoregressive generation, Tok directly measures generation length. We compare against Full Reasoning, Budget Force (BF), No-Think, and DEER; we also report S-GRPO, which requires additional LLM training.

\begin{table*}[t]
\centering
\resizebox{\textwidth}{!}{
\begin{tabular}{lrrrrrrrrrrrr}
\toprule
\multirow{2}{*}{\textbf{Method}} & \multicolumn{3}{c}{\textbf{AIME24}} & \multicolumn{3}{c}{\textbf{AIME25}} & \multicolumn{3}{c}{\textbf{GPQA-D}} & \multicolumn{3}{c}{\textbf{Average}}\\
\cmidrule(r){2-4} \cmidrule(r){5-7} \cmidrule(r){8-10} \cmidrule(r){11-13}
 & \textbf{Tok}$\downarrow$ {\scriptsize (\textbf{CR}\%$\downarrow$)} & \textbf{Acc}$\uparrow$ & \textbf{RPCR}$\uparrow$ & \textbf{Tok}$\downarrow$ {\scriptsize (\textbf{CR}\%$\downarrow$)} & \textbf{Acc}$\uparrow$ & \textbf{RPCR}$\uparrow$ & \textbf{Tok}$\downarrow$ {\scriptsize (\textbf{CR}\%$\downarrow$)} & \textbf{Acc}$\uparrow$ & \textbf{RPCR}$\uparrow$ & \textbf{Tok}$\downarrow$ {\scriptsize (\textbf{CR}\%$\downarrow$)} & \textbf{Acc}$\uparrow$ & \textbf{RPCR}$\uparrow$\\
\midrule
\rowcolor{gray!15}\multicolumn{13}{l}{\textbf{Based on Qwen3-8B}}\\
\midrule
Full & 15435 {\scriptsize (100\%)} & 72.22 & 33.84 & 17828 {\scriptsize (100\%)} & 63.33 & 23.68 & 9514 {\scriptsize (100\%)} & 60.10 & 23.09 & 14259 {\scriptsize (100\%)} & 65.22 & 27.22\\
No-Think & 7271 {\scriptsize (47\%)} & 19.99 & 0.00 & 5036 {\scriptsize (28\%)} & 21.11 & 0.00 & 2687 {\scriptsize (28\%)} & 38.13 & 0.00 &  4998{\scriptsize (35\%)} & 26.40 & 0.00\\
BF & 10373 {\scriptsize (67\%)} & 58.88 & 37.49 & 11772 {\scriptsize (66\%)} & 55.56 & 29.26 & 3962 {\scriptsize (42\%)} & 55.56 & 43.99 & 8702 {\scriptsize (61\%)} & 56.66 & 34.77\\
DEER & 13952 {\scriptsize (90\%)} & 72.22 & 37.44 & 16628 {\scriptsize (93\%)} & 67.78 & 28.07 & 9085 {\scriptsize (95\%)} & 59.60 & 23.63 & 13222 {\scriptsize (93\%)} & 66.53 & 30.35\\
RCPD (Ours)  & 9958 {\scriptsize (65\%)} & 72.22 & \textbf{52.45} & 10067 {\scriptsize (56\%)} & 63.33 & \textbf{41.94} & 4130 {\scriptsize (43\%)} & 64.65 & \textbf{64.21} & 8052 {\scriptsize (56\%)} & 66.73 & \textbf{50.09} \\
\hdashline 
S-GRPO* & 8810 {\scriptsize (57\%)} & 77.30 & 65.05 & - & - & - & 5271 {\scriptsize (55\%)} & 55.40 & 32.76 & - & - & -\\
\midrule
\rowcolor{gray!15}\multicolumn{13}{l}{\textbf{Based on Qwen3-14B}}\\
\midrule
Full & 13350 {\scriptsize (100\%)} & 75.56 & 35.79 & 16711 {\scriptsize (100\%)} & 70.00 & 27.26 & 7711 {\scriptsize (100\%)} & 64.31 & 13.81 & 12591 {\scriptsize (100\%)} & 69.96 & 27.54\\
No-Think & 6294 {\scriptsize (47\%)} & 27.78 & 0.00 & 3533 {\scriptsize (21\%)} & 24.44 & 0.00 & 2733 {\scriptsize (35\%)} & 53.66 & 0.00 & 4186 {\scriptsize (33\%)} & 35.29 & 0.00\\
BF & 8974 {\scriptsize (67\%)} & 54.44 & 29.71 & 10546 {\scriptsize (63\%)} & 56.67 & 30.56 & 3932 {\scriptsize (51\%)} & 62.29 & 21.95 & 7817 {\scriptsize (62\%)} & 57.80 & 28.80\\
DEER & 12265 {\scriptsize (92\%)} & 72.22 & 36.23 & 14387 {\scriptsize (86\%)} & 70.00 & 31.67 & 7274 {\scriptsize (94\%)} & 64.48 & 14.87 & 11309 {\scriptsize (90\%)} & 68.90 & 29.72\\
RCPD (Ours) & 8799 {\scriptsize (66\%)} & 73.33 & \textbf{51.77} & 9987 {\scriptsize (60\%)} & 68.89 & \textbf{44.51} & 3709 {\scriptsize (48\%)} & 67.17 & \textbf{36.42} & 7498 {\scriptsize (60\%)} & 69.80 & \textbf{46.03} \\
\hdashline 
S-GRPO* & 8932 {\scriptsize (67\%)} & 77.90 & 56.11 & - & - & - & 4537 {\scriptsize (59\%)} & 60.60 & 15.30 & - & - & -\\
\midrule
\rowcolor{gray!15} \multicolumn{13}{l}{\textbf{Based on Qwen3-30B-A3B}}\\
\midrule
Full & 13449 {\scriptsize (100\%)} & 82.22 & 41.30 & 16457 {\scriptsize (100\%)} & 74.44 & 31.06 & 7510 {\scriptsize (100\%)} & 66.50 & 25.17 & 12472 {\scriptsize (100\%)} & 74.38 & 33.56\\
No-Think & 5667 {\scriptsize (42\%)} & 26.67 & 0.00 & 3995 {\scriptsize (24\%)} & 23.33 & 0.00 & 2637 {\scriptsize (35\%)} & 47.60 & 0.00 & 4099 {\scriptsize (33\%)} & 32.53 & 0.00 \\
BF & 11147 {\scriptsize (83\%)} & 74.44 & 42.85 & 12708 {\scriptsize (77\%)} & 65.56 & 33.23 & 4617 {\scriptsize (61\%)} & 64.81 & 37.28 & 9490 {\scriptsize (76\%)} & 68.27 & 37.66\\
DEER & 12326 {\scriptsize (92\%)} & 80.00 & 43.27 & 15891 {\scriptsize (97\%)} & 67.78 & 27.97 & 7361 {\scriptsize (98\%)} & 67.85 & 27.51 & 11860 {\scriptsize (95\%)} & 71.87 & 33.17\\
RCPD (Ours) & 11229 {\scriptsize (83\%)} & 82.22 & \textbf{49.47} & 12670 {\scriptsize (77\%)} & 74.44 & \textbf{40.34} & 4469 {\scriptsize (60\%)} & 68.69 & \textbf{47.19} & 9457 {\scriptsize (76\%)} & 75.11 & \textbf{45.02}\\
\midrule
\rowcolor{gray!15} \multicolumn{13}{l}{\textbf{Based on DeepSeek-R1-Distill-8B}}\\
\midrule
Full & 21339 {\scriptsize (100\%)} & 76.67 & 26.56 & 23237 {\scriptsize (100\%)} & 70.00 & 21.04 & 11537 {\scriptsize (100\%)} & 58.73 & 17.86 & 18704 {\scriptsize (100\%)} & 68.47 & 22.49\\
BF   & 16202 {\scriptsize (76\%)} & 72.22 & 32.24 & 19740 {\scriptsize (85\%)} & 62.22 & 20.83 & 8909 {\scriptsize (77\%)} & 56.22 & 20.31 & 14950 {\scriptsize (80\%)} & 63.55 & 24.85\\
DEER & 21105 {\scriptsize (99\%)} & 72.22 & 24.75 & 23129 {\scriptsize (100\%)} & 67.77 & 20.17 & 11207 {\scriptsize (97\%)} & 58.25 & 17.95 & 18480 {\scriptsize (99\%)} & 66.08 & 21.47 \\
RCPD (Ours) & 16981 {\scriptsize (80\%)} & 76.67 & \textbf{33.38} & 19602 {\scriptsize (84\%)} & 67.77 & \textbf{23.80} & 9814 {\scriptsize (85\%)} & 58.25 & \textbf{20.50} & 15465 {\scriptsize (83\%)} & 67.56 & \textbf{26.61} \\
\bottomrule
\end{tabular}
}
\caption{Experimental results across various types of reasoning models. We categorize methods into inference-time (top, above dashed line) and training-based (bottom). S-GRPO is a training-based baseline. CR (relative to full decoding) is shown as an integer percentage in parentheses next to Tok, and we report RPCR (accuracy gain over the No-Think baseline per generated token) with two decimal places. DeepSeek-R1-Distill-8B lacks a No-Think mode; its RPCR is computed against the Qwen3-8B No-Think baseline. Gray rows indicate the training-based S-GRPO baseline. S-GRPO* indicates that the result comes directly from their paper, while ``-’’ signifies that the paper's results are missing.}
\label{tab:method_comparison}
\end{table*}

As shown in Table~\ref{tab:method_comparison}, results align with \textit{Reasoning Dynamics}, indicating that extending the thinking process into the Post-RCP convergence regime offers no accuracy gains. RCPD exploits this by substantially reducing token generation compared to full decoding while maintaining competitive performance. Crucially, as an inference-time method without fine-tuning, RCPD rivals the training-based S-GRPO baseline; specifically, it outperforms S-GRPO on GPQA-D despite underperforming on AIME24. This demonstrates that full decoding often involves redundant computation well beyond the necessary RCP.

To ensure a rigorous evaluation, we compare with the BF baseline under an iso-computational setting. BF exhibits significant degradation, underscoring the limitation of ignoring Reasoning Dynamics. By enforcing a static budget, BF often truncates the process during Pre-RCP, before the answer has converged. In contrast, RCPD adapts to instance-specific complexity, allowing reasoning to proceed until it enters Post-RCP convergence.

Finally, qualitative analyses further support the Post-RCP interpretation. Appendix~\ref{sec:appb} shows that once the model enters Post-RCP convergence, it often exhibits \textit{cognitive inertia}: subsequent tokens repeatedly rationalize the first-emerged answer rather than correcting it, and may even drift into redundant loops. RCPD truncates this regime by design, explaining why early stopping can improve both efficiency and robustness.

\section{Related Work}

Based on the efficient reasoning taxonomy established in~\cite{DBLP:journals/corr/abs-2503-16419,DBLP:journals/corr/abs-2503-24377}, we categorize related work into three classes, following and extending previous research: Post-training Based Methods, Prompt-based Methods, and Early Exit Methods.

\paragraph{Post-training Based Methods.}
These methods leverage supervised fine-tuning with variable-length chain-of-thought (CoT) data~\cite{DBLP:journals/corr/abs-2407-06023,DBLP:conf/aaai/KangSCZ25,DBLP:journals/corr/abs-2502-12067,DBLP:journals/corr/abs-2502-09601}, integrate length rewards in reinforcement learning~\cite{DBLP:journals/corr/abs-2501-12599,DBLP:journals/corr/abs-2501-12570,DBLP:journals/corr/abs-2503-04697}, or employ latent space reasoning methods~\cite{DBLP:journals/corr/abs-2412-06769,DBLP:journals/corr/abs-2412-13171,DBLP:journals/corr/abs-2502-21074}. While existing approaches reduce reasoning tokens through various training strategies, they require extensive data and retraining. In contrast, our method requires no training.

\paragraph{Prompt-based Methods.}
Approaches in this category~\cite{DBLP:journals/corr/abs-2412-18547,DBLP:journals/corr/abs-2501-19393,DBLP:journals/corr/abs-2503-01141,openai2025gptoss120bgptoss20bmodel} use different prompts to enforce the model to generate concise CoT, thus reducing unnecessary reasoning steps~\cite{DBLP:conf/aaai/KangSCZ25,DBLP:journals/corr/abs-2505-23480,DBLP:journals/corr/abs-2508-05337}. Although this approach has shown some success in generating more concise reasoning chains, it has significant limitations. Specifically, the quality and flexibility of the prompt design directly affect the model’s performance. In the face of diverse and complex reasoning tasks, static prompting methods often fail to adapt effectively to the variability of inputs, leading to reasoning chains that lack sufficient depth and precision.

\paragraph{Early Exit Methods.} These methods intervene in the reasoning process by prematurely terminating generation to mitigate redundancy. Existing approaches primarily fall into two categories. The first relies on external monitoring mechanisms or additional model training. For instance, methods utilizing auxiliary classifiers~\cite{liu-wang-2025-answer}, bandit controllers~\cite{DBLP:journals/corr/abs-2510-10103}, or difficulty estimators~\cite{DBLP:journals/corr/abs-2504-13367} require extra components, while \textbf{S-GRPO}~\cite{DBLP:journals/corr/abs-2505-07686} necessitates extra training of the LLM to learn early termination strategies. These approaches inevitably introduce additional training costs or inference latency. In contrast, a more recent line of work focuses on an overhead-free paradigm that leverages the model's intrinsic states without extra components. \textbf{DEER}~\cite{DBLP:journals/corr/abs-2504-15895} exemplifies this by enabling early exiting without computational overhead. Our study aligns with this efficient paradigm.
\section{Conclusion}
In this work, we reveal the Reasoning Dynamics of LLMs, characterized by Thinking-Content Compensation and Semantic Path Convergence. These dynamics shape the reasoning process into an active reasoning phase and a converged reasoning phase separated by the RCP. We demonstrate that a simple method can effectively detect this point to mitigate overthinking. Our findings identify post-RCP redundancy as the primary cause of overthinking, highlighting reliable self-termination as a cornerstone for efficient test-time scaling.

\section{Limitations}
Our analysis and evaluation focus on reasoning models and are conducted primarily on the Qwen3 and DeepSeek-R1 families, spanning multiple model sizes and benchmarks. We prioritize these open-weight systems because they expose explicit thinking delimiters and the decoding signals needed for controlled interventions and RCPD. While these models exhibit consistent Reasoning Dynamics, we did not run the full suite of RCP analyses across the broader ecosystem of models, such as closed-source APIs, tool-augmented agents and multimodal models. Establishing how universal RCP-like phase transitions are beyond this scope remains an important direction.

\section{Potential Risks}
The proposed RCPD method primarily focuses on enhancing the efficiency of reasoning processes in LLMs. As such, there are no inherent risks associated with its implementation or deployment. The method does not introduce any new computational mechanisms that could adversely affect the model’s operation. Additionally, RCPD aims to mitigate overthinking in LLMs, which, in turn, reduces unnecessary resource consumption and computational overhead.

\bibliography{custom}

\newpage

\appendix

\section{Additional Experiment}
Due to space constraints, we report the Qwen3-32B results in Table~\ref{tab:app_method_comparison} in the Appendix.

\section{Additional Evaluation Details}
\label{sec:appendix_eval_details}
This section complements the main experimental description by specifying the evaluation protocol and baseline implementations used for early-exit comparisons. Specifically, our evaluation utilizes AIME24 and AIME25, which each contain 30 problems, and GPQA-D, which comprises 198 problems.

\paragraph{Decoding Configuration and Token Accounting.}
We use each model's official chat template and its delimiter-based reasoning format (e.g., \texttt{<think>...\ </think>}).
Unless otherwise stated, we follow the recommended nucleus-sampling settings (Temperature $=0.6$, Top-$p=0.95$) and set the maximum generation length to 32,768 tokens so that behaviors are not driven by hard token caps.

We report \textbf{Tok} as the average number of generated tokens per instance (including both thinking and content). To measure the efficiency trade-off, we report the relative performance--cost ratio (\textbf{RPCR}), defined as the accuracy gain over the No-Think baseline per generated token:
\[
\text{RPCR} = \frac{\text{Acc} - \text{Acc}_{\text{No-Think}}}{\text{Tok}} \times 10^{4}
\]
For readability, we report \textbf{RPCR} with two decimal places. Additionally, the Compression Rate (\textbf{CR}), defined as the token ratio relative to full decoding (i.e., $\mathrm{Tok}/\mathrm{Tok}_{\text{Full}}$), is shown as an integer percentage in parentheses next to \textbf{Tok}.

\paragraph{Baselines.}
We compare RCPD against representative inference-time strategies:
\begin{itemize}
    \item \textbf{Full:} standard decoding without early stopping.
    \item \textbf{No-Think:} prompting the model to skip explicit reasoning and answer directly.
    \item \textbf{BF (Budget Force):} static truncation that forces \texttt{</think>} after a fixed thinking budget.
    \item \textbf{DEER:} a trigger-based early-exit method that terminates upon reflective cue words and a confidence threshold on the final answer format.
    \item \textbf{S-GRPO:} a training-based early-exit policy (we report results directly from the original paper when available).
\end{itemize}

\paragraph{Iso-Compute Calibration for BF.}
To make BF a strong baseline, we evaluate it under an oracle iso-computational setting: for each (model, benchmark), we retrospectively select the BF budget so that its average token usage matches RCPD.
This gives BF the advantage of using a globally tuned budget that is not available at deployment time.

\begin{figure*}[t]                     
  \centering
  \begin{subfigure}[t]{0.48\textwidth}
    \centering
    \includegraphics[width=0.90\linewidth]{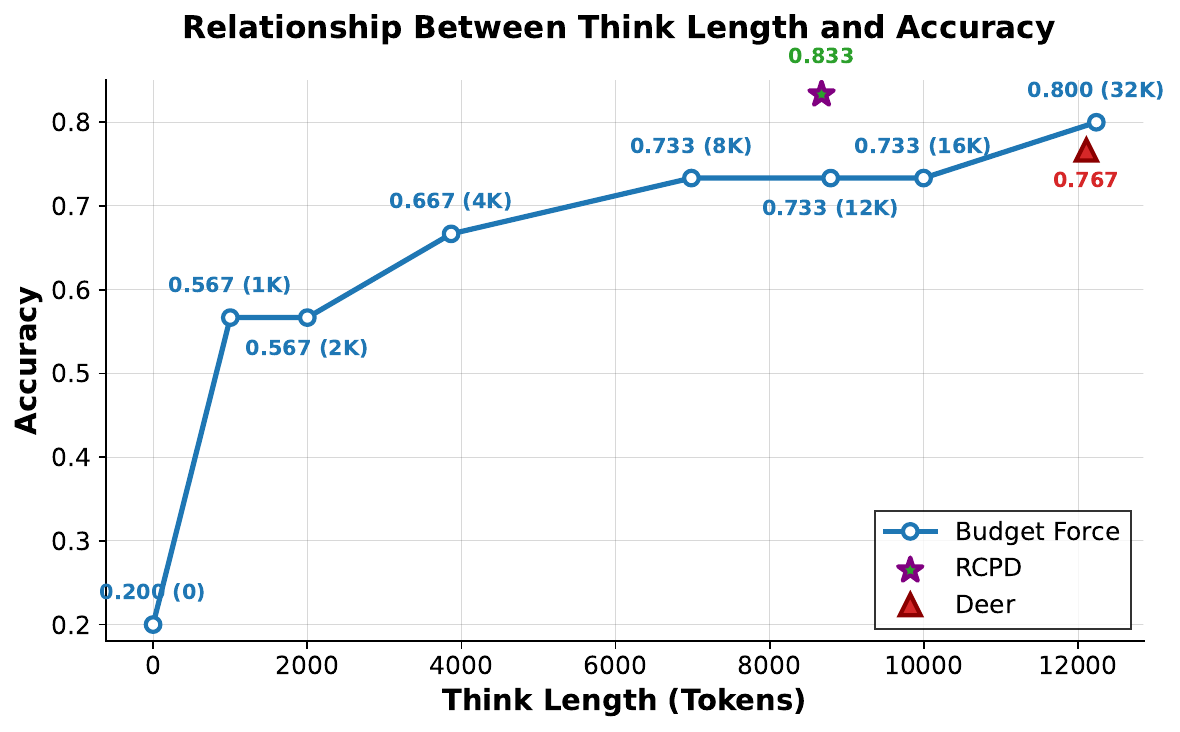}
    \caption{Relationship between thinking length and accuracy}
    \label{fig:bfA}
  \end{subfigure}
  \hfill
  \begin{subfigure}[t]{0.48\textwidth}
    \centering
    \includegraphics[width=0.90\linewidth]{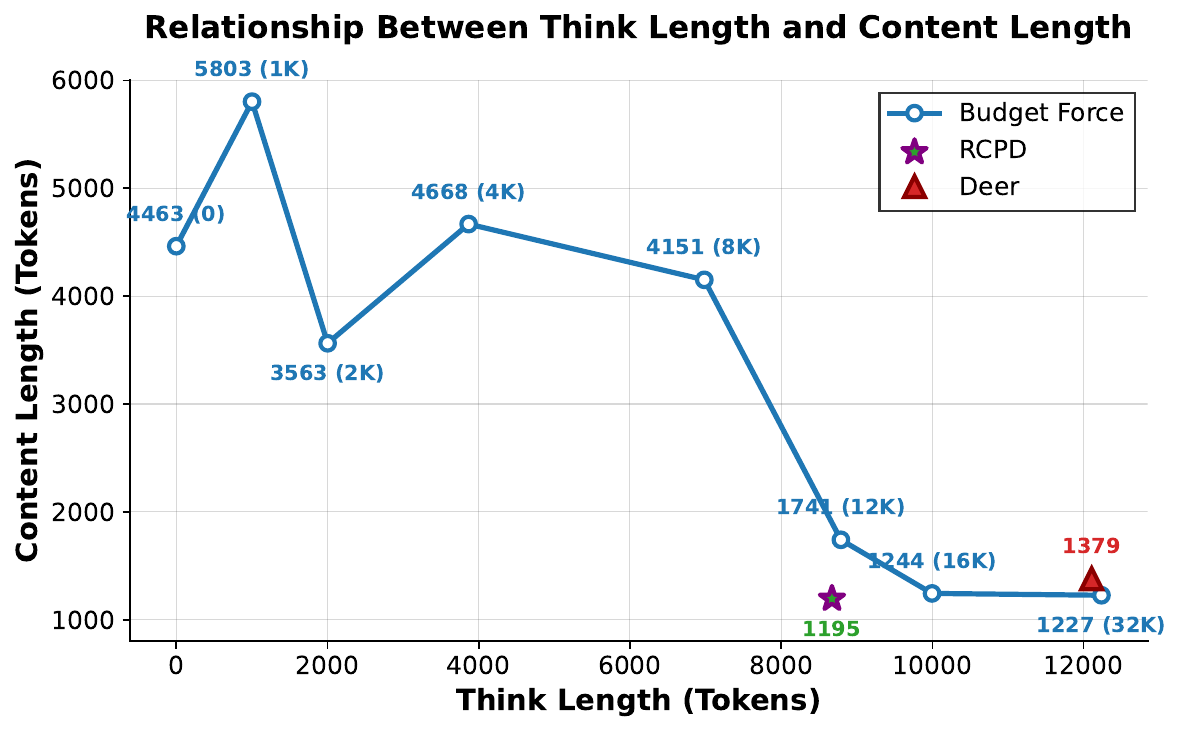}
    \caption{Relationship between thinking length and content length}
    \label{fig:bfB}
  \end{subfigure}
  \caption{Static truncation (Budget Force) cannot reliably trade compute for accuracy, while RCPD stops near the saturation point and reduces tokens with minimal accuracy loss.}
  \label{fig:bf}
\end{figure*}

\paragraph{From Static Budgets to Adaptive Stopping.}
Figure~\ref{fig:bf} visualizes a BF sweep on AIME24 (Qwen3-32B).
Accuracy improves with more thinking only up to a narrow saturation region, while token usage continues to grow.
Consequently, a single fixed budget cannot reliably stop near the saturation point across instances, whereas RCPD adapts to online convergence signals and thus achieves a more consistent compute--accuracy trade-off.

\paragraph{Impact of Temporal Consistency on Stopping Robustness.}
Table~\ref{tab:rcpd_ablation} compares RCPD with a simplified baseline that triggers stopping as soon as the \texttt{</think>} token enters the top-5 of the next-token distribution at a sentence boundary (denoted as \texttt{</think>}-5).
While this single-threshold rule confirms that delimiter rank is a meaningful convergence proxy, RCPD performs better by incorporating short-horizon temporal patterns (Appendix~\ref{sec:appendix_stopping}) that reduce sensitivity to noise.

\section{Implementation Details of RCPD Stopping Criteria}
\label{sec:appendix_stopping}

As discussed in Section \S3.3, to balance computational efficiency with detection accuracy, we distill the decision boundaries mined from the probe dataset into a set of deterministic, symbolic rules. The generation process terminates immediately if the rank trajectory of the \texttt{</think>} token, denoted as $R_t$, satisfies any of the following logical conditions.

To facilitate reproducibility and future research, we will make our code and data publicly available upon acceptance.

\paragraph{1. Immediate Saturation.}
This condition captures scenarios where the model exhibits overwhelming certainty. A rank of $R_t \le 5$ indicates that the reasoning process has effectively collapsed into a definitive conclusion, rendering further generation redundant.
\begin{equation}
    \mathcal{C}_{\text{sat}}: R_t \le 5
\end{equation}

\paragraph{2. Progressive Convergence.}
This rule corresponds to the ``ladder pattern'' observed in successful deductions. It requires a monotonic ascent in the token's rank probability (i.e., descending rank values) over a fixed window, representing a directed convergence toward the solution.
\begin{equation}
    \begin{split}
        \mathcal{C}_{\text{conv}}: \bigwedge_{k=0}^{3} (R_{t-k} \le \theta_k), \\
        \text{where } \boldsymbol{\theta} = [10, 50, 100, 1000]
    \end{split}
\end{equation}

\paragraph{3. Sustained Plateau.}
To distinguish between true convergence and stochastic noise, this condition detects whether the model has reached an ``entropic floor.'' Even if the rank does not hit the absolute peak ($R_t \approx 1$), a sustained presence within a low-rank region signals that no new information is being generated. We define this via two sub-conditions:

\begin{itemize}
    \item \textbf{Short-term Stability (Noise Filtering):} Ensures high confidence is maintained over a narrow window.
    \begin{equation}
        \mathcal{C}_{\text{short}}: \forall k \in \{0, 1, 2\}, R_{t-k} \le 20
    \end{equation}
    
    \item \textbf{Long-term Persistence (Sampling Inertia):} Detects a prolonged plateau where the model effectively loops or stalls in a high-probability region.
    \begin{equation}
        \mathcal{C}_{\text{long}}: \forall k \in \{0, \dots, 5\}, R_{t-k} \le 50
    \end{equation}
\end{itemize}

The final stop signal $S_t$ is triggered by the disjunction of these conditions:
\begin{equation}
    S_t = \mathcal{C}_{\text{sat}} \lor \mathcal{C}_{\text{conv}} \lor (\mathcal{C}_{\text{short}} \lor \mathcal{C}_{\text{long}})
\end{equation}

\section{Quantifying Semantic Convergence}
\label{sec:appendix_semantic}
We quantify the stability of the semantic trajectory after the RCP by fitting a 95\% confidence ellipse in the 2D PCA space (Figure~\ref{fig:semantic_trajectory}). Let $\hat{k}$ denote the RCP step, and let $\tilde{\mathbf{v}}_{k}^{(i)}$ denote the PCA-projected embedding of the $i$-th sampled continuation at step $k$. We collect all post-RCP points into $\mathcal{C} = \{\tilde{\mathbf{v}}_{k}^{(i)} : k \ge \hat{k}\}$ and estimate their sample mean $\boldsymbol{\mu}$ and covariance $\Sigma$.

For any point $\mathbf{x}$ in this PCA plane, we measure its semantic deviation from the converged region using the squared Mahalanobis distance:
\begin{equation}
    D_M(\mathbf{x})^2 = (\mathbf{x} - \boldsymbol{\mu})^T \Sigma^{-1} (\mathbf{x} - \boldsymbol{\mu})
    \label{eq:mahalanobis}
\end{equation}
We then define the 95\% confidence ellipse (semantic basin) as the level set:
\begin{equation}
    \mathcal{B}_{0.95} = \left\{\mathbf{x} \mid D_M(\mathbf{x})^2 \le \chi^2_2(0.95)\right\},
    \label{eq:confidence_ellipse}
\end{equation}
where $\chi^2_2(0.95) \approx 5.991$ is the 95th percentile of the $\chi^2$ distribution with 2 degrees of freedom. Under a locally elliptical (approximately Gaussian) stationary regime in the PCA plane, $D_M(\mathbf{x})^2$ follows $\chi^2_2$, so $\mathcal{B}_{0.95}$ contains 95\% probability mass of the converged distribution.

Finally, we compute the fraction of post-RCP points that remain within this basin. In Figure~\ref{fig:semantic_trajectory}, 97.3\% of subsequent trajectory points lie inside $\mathcal{B}_{0.95}$, indicating that after the RCP the semantic path no longer drifts but fluctuates within a compact, steady neighborhood.

\begin{table}[t]
\centering
\resizebox{0.45\textwidth}{!}{
\begin{tabular}{c c c}
\hline
\textbf{Model} & \textbf{Method} & \textbf{Acc} \\
\hline
Qwen3-8B & \texttt{</think>}-5 & 61.11 \\
Qwen3-8B & RCPD & 75.11 \\
\midrule
Qwen3-14B & \texttt{</think>}-5 & 65.72 \\
Qwen3-14B & RCPD & 69.79 \\
\midrule
Qwen3-30B-A3B & \texttt{</think>}-5 & 73.84 \\
Qwen3-30B-A3B & RCPD & 75.11 \\
\midrule
Qwen3-32B & \texttt{</think>}-5 & 73.26 \\
Qwen3-32B & RCPD & 73.37 \\
\hline
\end{tabular}
}
\caption{Static stopping by a single rank threshold (\texttt{</think>}-5) is competitive but consistently worse than RCPD, indicating that simple temporal patterns improve robustness.}
\label{tab:rcpd_ablation}
\end{table}

\section{Additional Visualizations}
\label{sec:appendix_additional_visualizations}
To demonstrate that the same phenomena consistently appear across datasets, we provide additional examples that could not fit in the main text.

\begin{figure*}[t]
  \centering
  \begin{subfigure}[t]{0.48\textwidth}
    \centering
    \includegraphics[width=0.9\linewidth]{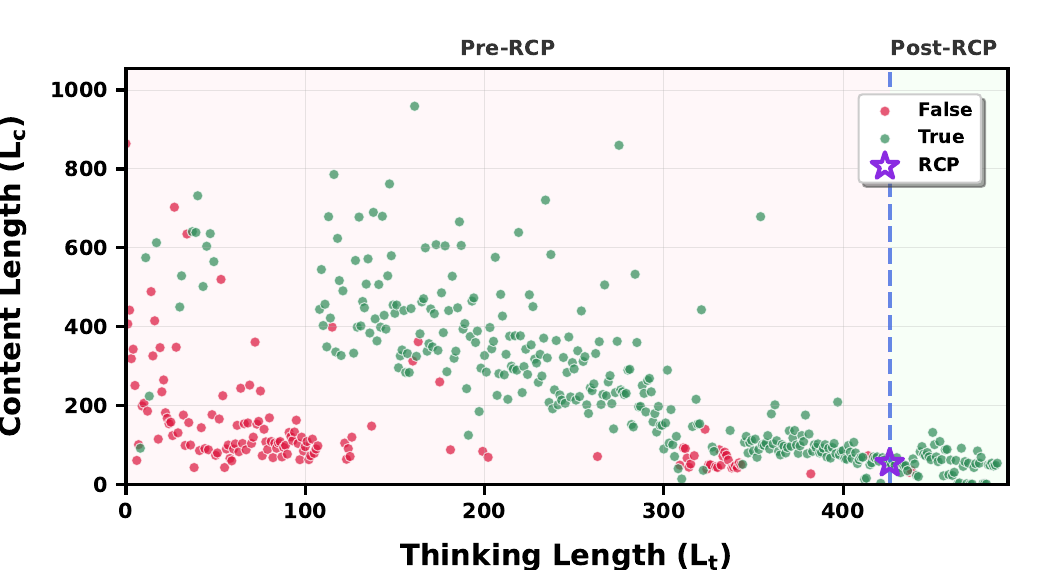}
    \caption{AIME24 Question 2 Length Dynamics}
    \label{fig:appendix_three_stage_aime24_29}
  \end{subfigure}
  \hfill
  \begin{subfigure}[t]{0.48\textwidth}
    \centering
    \includegraphics[width=0.9\linewidth]{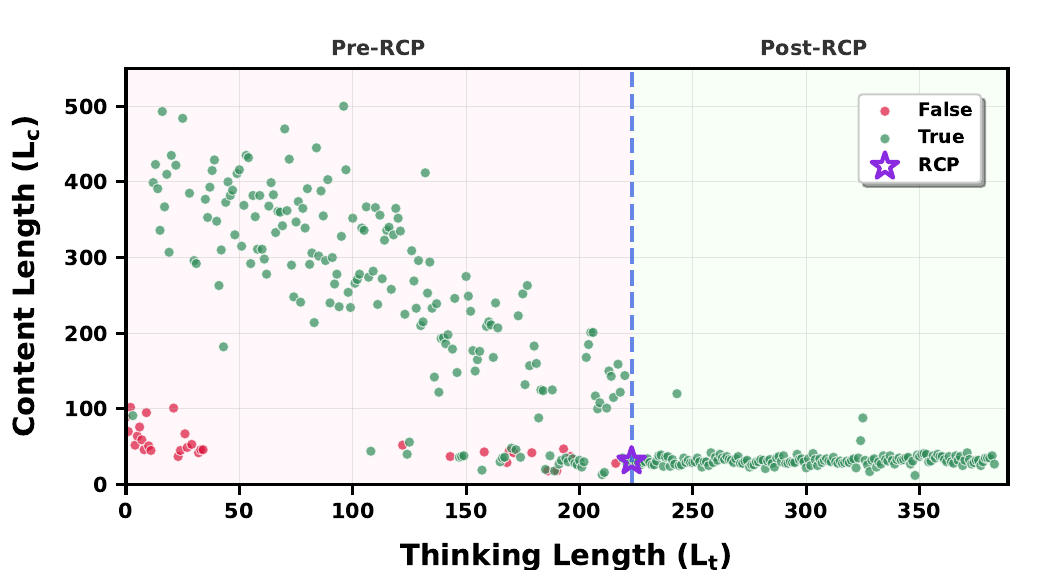}
    \caption{AIME24 Question 9 Length Dynamics}
    \label{fig:appendix_three_stage_gpqa_15}
  \end{subfigure}
  \caption{Supplementary examples illustrating the transition from the Pre-RCP Active Reasoning Stage to the Post-RCP Converged Reasoning Stage (complementing Figure~\ref{fig:two_three_stage}).}
  \label{fig:appendix_three_stage_additional}
\end{figure*}

\begin{figure*}[t]
    \centering
    \begin{subfigure}[t]{0.48\textwidth}
        \centering
        \includegraphics[width=\linewidth]{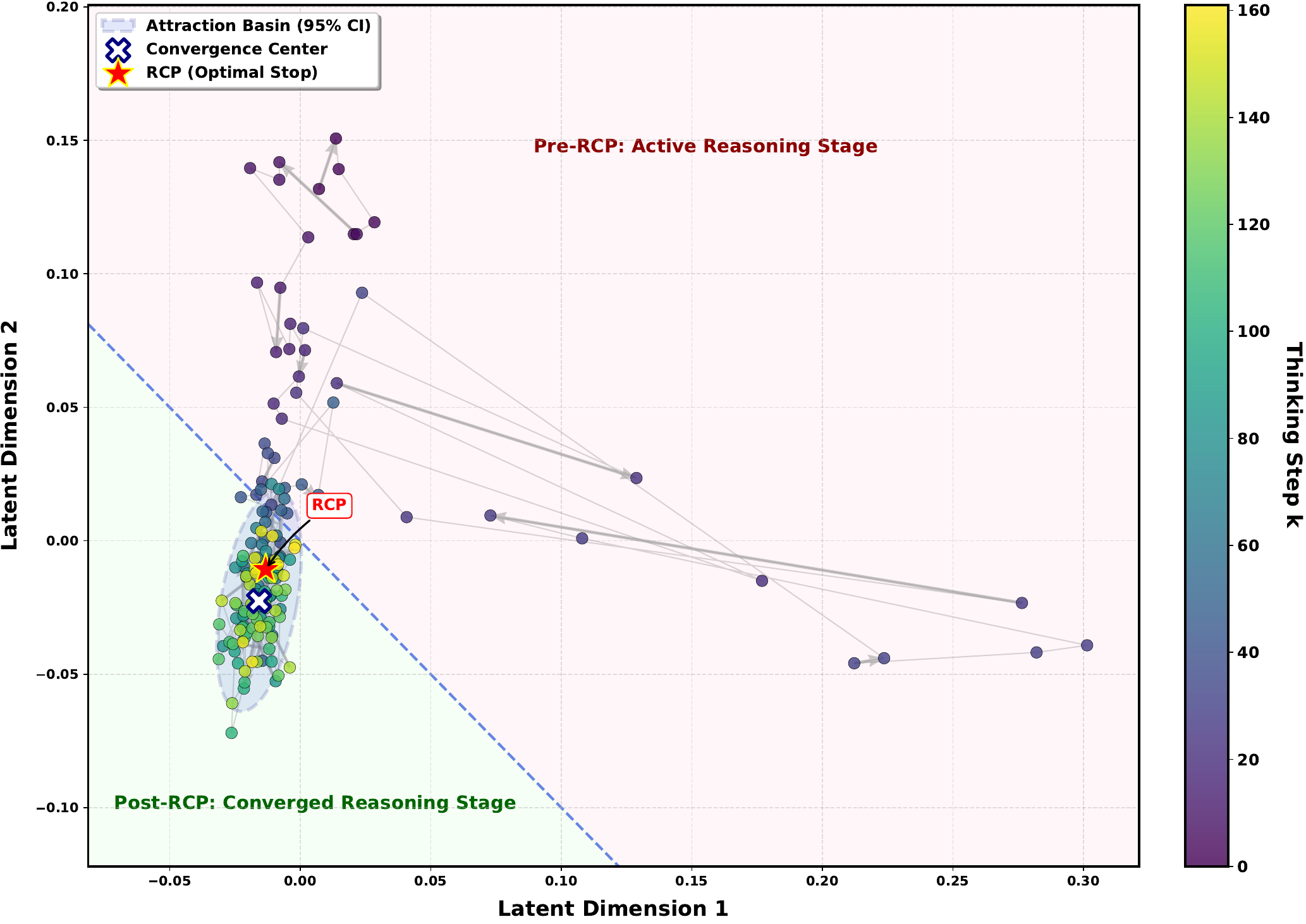}
        \caption{AIME24 Question 1 Semantic Trajectory}
        \label{fig:appendix_semantic_traj_aime24_1}
    \end{subfigure}
    \hfill
    \begin{subfigure}[t]{0.48\textwidth}
        \centering
        \includegraphics[width=\linewidth]{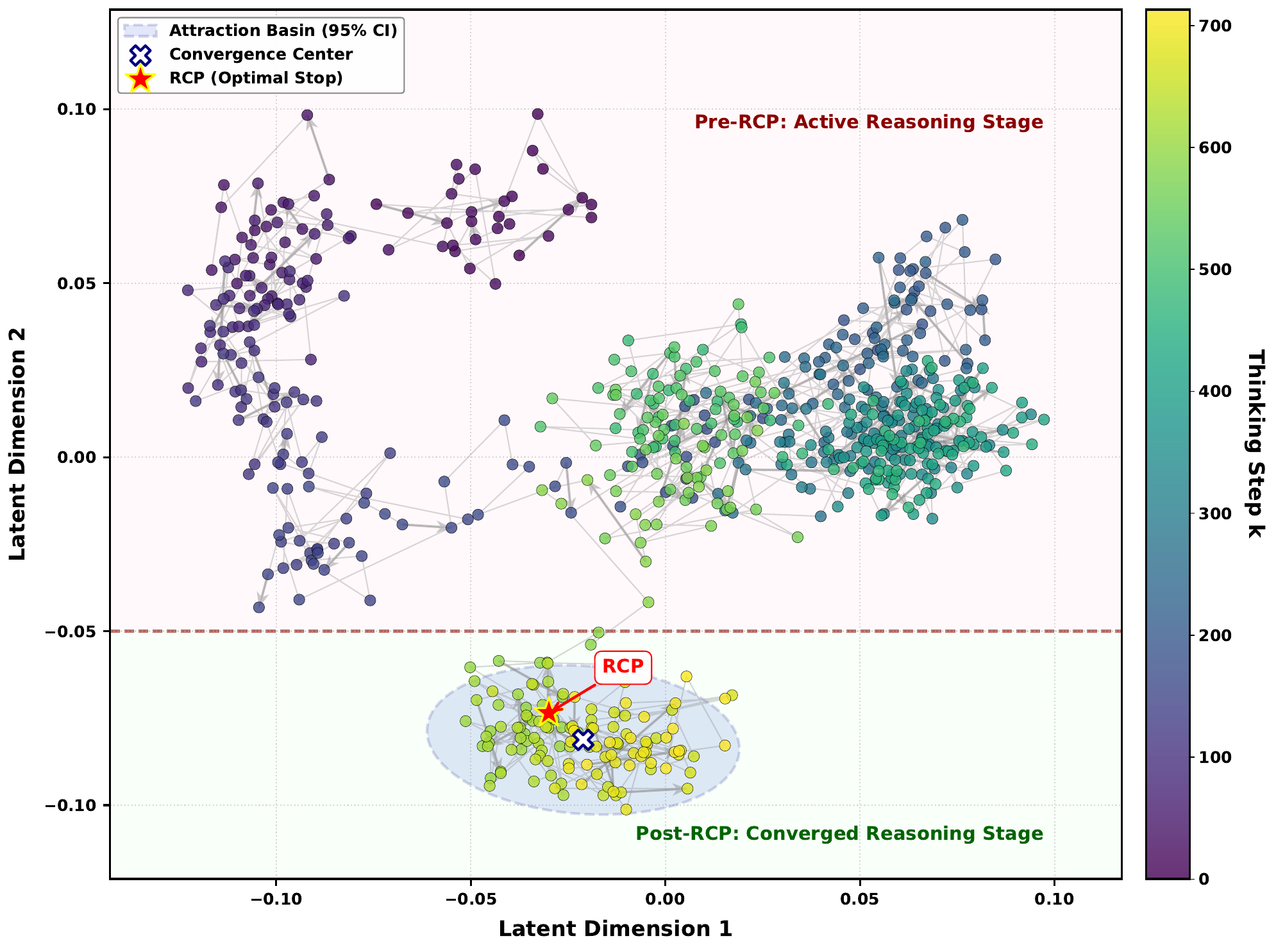}
        \caption{AIME24 Question 30 Semantic Trajectory}
        \label{fig:appendix_semantic_traj_aime24_30}
    \end{subfigure}
    \caption{Additional semantic trajectory examples (supplementary to Figure~\ref{fig:semantic_trajectory}).}
    \label{fig:appendix_semantic_trajectory_additional}
\end{figure*}

Using the same 95\% confidence ellipse protocol in Section~\ref{sec:appendix_semantic}, we quantify post-RCP semantic stability for these additional cases. For AIME24 Question 1 (Figure~\ref{fig:appendix_semantic_traj_aime24_1}), 94.3\% of Post-RCP points lie inside the 95\% basin, and for AIME24 Question 30 (Figure~\ref{fig:appendix_semantic_traj_aime24_30}), the ratio is 96.5\%. The consistently high in-basin fractions support that semantic convergence after the RCP is not instance-specific but recurs across different problems.

\begin{table*}
\centering
\setlength{\tabcolsep}{12pt}
\begin{tabular}{l c c c}
\toprule
\multirow{2}{*}{\textbf{Experimental Setting}} & \multicolumn{3}{c}{\textbf{Final Answer Distribution (out of 64)}} \\
\cmidrule(lr){2-4}
 & \textbf{Recovered Original} & \textbf{Adopted Injection} & \textbf{Other} \\
\midrule
Injection: \textbf{16} (Original: 12) & 11 & \textbf{49} & 4 \\

Injection: \textbf{12} (Original: 16) & 11 & \textbf{51} & 2 \\
\bottomrule
\end{tabular}
\caption{Impact of Modified Intermediate Answer on Final Reasoning Outcome. The results demonstrate a strong confirmation bias: in both settings, the model overwhelmingly adopts the injected intermediate answer rather than recovering the correct original answer.}
\label{tab:false_injection}
\end{table*}

\begin{table*}
\centering
\resizebox{\textwidth}{!}{
\begin{tabular}{lrrrrrrrrrrrr}
\toprule
\multirow{2}{*}{\textbf{Method}} & \multicolumn{3}{c}{\textbf{AIME24}} & \multicolumn{3}{c}{\textbf{AIME25}} & \multicolumn{3}{c}{\textbf{GPQA-D}} & \multicolumn{3}{c}{\textbf{Average}}\\
\cmidrule(r){2-4} \cmidrule(r){5-7} \cmidrule(r){8-10} \cmidrule(r){11-13}
 & \textbf{Tok}$\downarrow$ {\scriptsize (\textbf{CR}\%$\downarrow$)} & \textbf{Acc}$\uparrow$ & \textbf{RPCR}$\uparrow$ & \textbf{Tok}$\downarrow$ {\scriptsize (\textbf{CR}\%$\downarrow$)} & \textbf{Acc}$\uparrow$ & \textbf{RPCR}$\uparrow$ & \textbf{Tok}$\downarrow$ {\scriptsize (\textbf{CR}\%$\downarrow$)} & \textbf{Acc}$\uparrow$ & \textbf{RPCR}$\uparrow$ & \textbf{Tok}$\downarrow$ {\scriptsize (\textbf{CR}\%$\downarrow$)} & \textbf{Acc}$\uparrow$ & \textbf{RPCR}$\uparrow$\\
\midrule
\rowcolor{gray!15} \multicolumn{13}{l}{\textbf{Based on Qwen3-32B}}\\
\midrule
Full & 11955 {\scriptsize (100\%)} & 82.22 & 40.90 & 16878 {\scriptsize (100\%)} & 65.56 & 28.31 & 7357 {\scriptsize (100\%)} & 69.53 & 25.35 & 12063 {\scriptsize (100\%)} & 72.44 & 31.87\\
No-Think & 5731 {\scriptsize (48\%)} & 33.33 & 0.00 & 5472 {\scriptsize (32\%)} & 17.78 & 0.00 & 2277 {\scriptsize (31\%)} & 50.88 & 0.00 & 4493 {\scriptsize (37\%)} & 33.99 & 0.00\\
BF   & 10071 {\scriptsize (84\%)} & 78.89 & 45.24 & 11772 {\scriptsize (70\%)} & 55.56 & 32.09 & 4137 {\scriptsize (56\%)} & 64.98 & 34.08 & 8660 {\scriptsize (72\%)} & 66.48 & 37.52\\
DEER & 12002 {\scriptsize (100\%)} & 81.11 & 39.81 & 14905 {\scriptsize (88\%)} & 66.67 & 32.80 & 6841 {\scriptsize (93\%)} & 69.70 & 27.51 & 11249 {\scriptsize (93\%)} & 72.49 & 34.23\\
RCPD  & 10062 {\scriptsize (84\%)} & 82.22 & \textbf{48.59} & 10917 {\scriptsize (65\%)} & 66.67 & \textbf{44.78} & 4094 {\scriptsize (56\%)} & 71.21 & \textbf{49.66} & 8358 {\scriptsize (69\%)} & 73.37 & \textbf{47.12}\\
\bottomrule
\end{tabular}
}
\caption{Additional results on Qwen3-32B, reporting Tok (CR\%), Acc, and $10^{4}\times$RPCR (two decimals).}
\label{tab:app_method_comparison}
\end{table*}

\section{Examples of Overthinking}
\label{sec:appb}
This section describes three typical cases of overthinking phenomena in model reasoning. The first case involves repeated verification of a correct answer, while the other two cases involve repeated derivation of speculative incorrect answers. These examples illustrate how overthinking can lead to unnecessary computational waste and inefficient problem-solving processes.

To further assess the effectiveness of the model's post-RCP validation mechanism, we conduct a counterfactual modification experiment. We manually edit the initial tentative answer in the reasoning trace of Table~\ref{tab:false_injection}. Concretely, we replace ``\ldots perhaps the number of intersections is 12\ldots'' with ``\ldots perhaps the number of intersections is 16\ldots''. After injecting this incorrect intermediate conclusion, we run 64 independent reasoning trials. As shown in Table~\ref{tab:false_injection}, the model exhibits a pronounced \textbf{confirmation bias}. Only 11 trials recover the correct answer (12), whereas 49 out of 64 converge to the injected value (16). We observe the same pattern under the symmetric setting that flips the injection from 16 to 12, where the model again predominantly adopts the newly specified answer. Overall, these results indicate that Post-RCP generation seldom performs genuine re-verification and instead tends to rationalize earlier tokens.

\begin{table*}
    \centering
    \small
    \noindent\fbox{%
    \begin{minipage}{2.0\columnwidth} 
\tt 

User:  Please reason step by step, and put your final answer within \\boxed{}. Find the number of rectangles that can be formed inside a fixed regular dodecagon ($12$-gon) where each side of the rectangle lies on either a side or a diagonal of the dodecagon. The diagram below shows three of those rectangles.

Assistant: ... Therefore, returning to the dodecagon, with 6 diameters, the number of rectangles would be C(6,2)=15. But the problem says that the diagram shows three of those rectangles, and asks for the total number. Maybe 15 is the answer?\newline
... Therefore, our formula seems valid. Therefore, returning to the dodecagon. If there are 6 diameters, then number of rectangles is C(6,2) = 15. But the problem says "three of those rectangles" are shown in the diagram. Maybe the diagram is just an example, and the answer is indeed 15? \newline
... For example, choosing 0-6 and 1-7 gives a rectangle with vertices 0,1,6,7. Choosing 0-6 and 2-8 gives rectangle 0,2,6,8. Etc. Each of these should be unique. Therefore, total number is 15. \newline
... Therefore, the rectangle must be inscribed in the same circle. Therefore, must have diagonals as diameters. Therefore, 15. \newline
... \newline
Therefore, I think the answer is 15.

**Final Answer**\\
\boxed{15}
\end{minipage}
}
\caption{As shown in the table, the answer 15 was repeatedly derived through a speculative combinatorial formula during the problem-solving process. The assistant multiple times applied the same formula \( \binom{6}{2} = 15 \) based on the assumption of 6 diameters, without rigorous validation or alternative approaches. For conciseness, only representative instances of this repetitive reasoning are included. This circular derivation process illustrates a clear case of overthinking, resulting in unnecessary computational effort.}
\label{tab:false_2}
\end{table*}

\begin{table*}
    \centering
    \small
    \noindent\fbox{%
    \begin{minipage}{2.0\columnwidth} 
\tt 

User: Please reason step by step, and put your final answer within \texttt{\boxed{}}. Every morning Aya goes for a $9$-kilometer-long walk and stops at a coffee shop afterwards. When she walks at a constant speed of $s$ kilometers per hour, the walk takes her 4 hours, including $t$ minutes spent in the coffee shop. When she walks $s+2$ kilometers per hour, the walk takes her 2 hours and 24 minutes, including $t$ minutes spent in the coffee shop. Suppose Aya walks at $s+\frac{1}{2}$ kilometers per hour. Find the number of minutes the walk takes her, including the $t$ minutes spent in the coffee shop. 

Assistant: ... The walking time is 3 hours, coffee shop time is 24 minutes. So total time is 3 hours + 24 minutes. Let me convert 3 hours to minutes: 180 minutes + 24 minutes = 204 minutes. \newline
... So 204 minutes.\newline
Wait, but let me check again. Wait, 3 hours is 180 minutes. 180 + 24 = 204. \newline
So 204 minutes. Is that the answer?\newline
... \newline
Then adding 24 minutes gives 204 minutes. Therefore, the answer is 204. \newline
Alternatively, convert 24 minutes to 0.4 hours, total time 3.4 hours, which is 3 hours 24 minutes. Either way, the problem asks for the number of minutes. So 204 minutes.\\

**Final Answer**\\
\boxed{204}
\end{minipage}
}
\caption{As shown in the table, the answer 204 was repeatedly verified multiple times during the problem-solving process. The reasoning process involved 20 instances of repeated verification, and for the sake of brevity, only the most representative instances are included in the table. This repetitive verification process led to a significant waste of computational resources, which is a typical case of overthinking.}
\label{tab:true}
\end{table*}

\begin{table*}
    \centering
    \small
    \noindent\fbox{%
    \begin{minipage}{2.0\columnwidth} 
\tt 

User:  Please reason step by step, and put your final answer within \boxed{}. Define $f(x)=|| x|-\tfrac{1}{2}|$ and $g(x)=|| x|-\tfrac{1}{4}|$. Find the number of intersections of the graphs of $[y=4 g(f(\sin (2 \pi x))) \quad\text{ and }\quad x=4 g(f(\cos (3 \pi y))).]$

Assistant: ... This is a complex equation, but perhaps we can analyze it by considering the possible values. Let me note that both x and y are in [0,1], and the functions are continuous. Therefore, the number of solutions is finite, and we can try to find them by considering the number of times the curves cross. Alternatively, since both functions are symmetric and periodic, perhaps the number of intersections is 12, or some other number. \newline
... let me consider the following: each time the function F(x) completes an oscillation, it crosses the function G(y) multiple times. Since F(x) has two oscillations and G(y) has three, the total number of intersections could be 2 * 3 * 2 = 12, but this is just a guess. \newline
... Given the complexity, perhaps the number of intersections is 12, as each oscillation of F(x) (2 oscillations) intersects each oscillation of G(y) (3 oscillations) twice, giving 2 * 3 * 2 = 12. \newline
... Given the complexity, it's hard to say, but if each oscillation intersects twice, then 2 * 3 * 2 = 12 might be the answer. \newline
... \newline
Given that, and considering that (1,1) is one of them, I'll go with 12 as the answer.

**Final Answer**\\
\boxed{12}
\end{minipage}
}
\caption{As shown in the table, the answer 12 was repeatedly speculated multiple times during the problem-solving process. The reasoning involved several instances of similar speculation based on oscillatory patterns and multiplicative guesses, and for brevity, only representative excerpts are included. This repetitive speculation process led to a waste of computational resources, exemplifying overthinking.}
\label{tab:false_1}
\end{table*}

\section{Declaration of AI Use}
\label{sec:ai_declaration}

In accordance with the ACL Policy on AI Writing Assistance, we openly disclose the use of AI tools in the preparation of this manuscript.

\paragraph{Writing Assistance}
We utilized Gemini 3-Pro solely as a language editing tool. Its application was strictly limited to paraphrasing, checking grammar, and polishing the original text written by the authors to enhance readability and clarity. The AI tool was not used to generate new scientific concepts, formulate hypotheses, or interpret the results. The authors have reviewed all AI-assisted text and retain full responsibility for the accuracy, originality, and integrity of the content.

\paragraph{Experimental Models}
It is important to distinguish the writing assistant from the models studied in this work. The methodologies, experiments, and results presented in this paper are based on the \textbf{Qwen3} and \textbf{DeepSeek-R1} large language models. These models served exclusively as the subjects of our research and the foundation for our experimental framework; they were not involved in the writing or drafting of this manuscript.

\end{document}